%% file: acl_latex.tex
\pdfoutput=1

\documentclass[11pt]{article}

\usepackage[final]{acl}

\usepackage{times}
\usepackage{latexsym}
\usepackage[T1]{fontenc}
\usepackage[utf8]{inputenc}
\usepackage{microtype}
\usepackage{inconsolata}
\usepackage{graphicx}

\usepackage[utf8]{inputenc} 
\usepackage[T1]{fontenc}    
\usepackage{hyperref}       
\usepackage{url}            
\usepackage{booktabs}       
\usepackage{amsfonts}       
\usepackage{nicefrac}       
\usepackage{microtype}      
\usepackage{xcolor}         

\usepackage{multirow}
\usepackage{subfigure}

\usepackage[english]{babel}
\usepackage{moresize}
\usepackage{amsmath}
\usepackage{algorithmic}
\usepackage{balance}
\usepackage{comment}
\usepackage{paralist}
\usepackage{bm}
\usepackage{pgfplots}
\usetikzlibrary{pgfplots.dateplot}

\usepackage{flushend}
\usepackage[english]{babel}
\usepackage{graphicx}

\usepackage{amssymb}
\usepackage{amsfonts}
\usepackage{url}
\usepackage{bbm}
\usepackage{longtable}
\usepackage{rotating}
\usepackage{multirow}
\usepackage{mathrsfs}
\usepackage{enumitem}
\usepackage[linesnumbered,algoruled,boxed,lined]{algorithm2e}
\usepackage{adjustbox}
\usepackage{hyperref}
\usepackage{pgfplots}
\usetikzlibrary{pgfplots.dateplot}
\usepackage{filecontents}

\newcommand{\eg}{\textit{e}.\textit{g}.}

\def\model{OpenGraph}

\newcommand{\graph}{\mathcal{G}}
\newcommand{\setv}{\mathcal{V}}
\newcommand{\sete}{\mathcal{E}}

\newcommand{\setc}{\mathcal{C}}
\newcommand{\setp}{\mathcal{P}}
\newcommand{\complexity}{\mathcal{O}}
\newcommand{\matf}{\textbf{F}}
\newcommand{\dmnr}{\mathbb{R}}
\newcommand{\veca}{\textbf{a}}
\newcommand{\vece}{\textbf{e}}
\newcommand{\vech}{\textbf{h}}
\newcommand{\mata}{\textbf{A}}
\newcommand{\matd}{\textbf{D}}

\newcommand{\matu}{\textbf{U}}
\newcommand{\matv}{\textbf{V}}
\newcommand{\matw}{\textbf{W}}
\newcommand{\loss}{\mathcal{L}}
\newcommand{\param}{\mathbf{\Theta}}

\title{OpenGraph: Towards Open Graph Foundation Models}

\author{Lianghao Xia \\
  University of Hong Kong \\
  \texttt{aka\_xia@foxmail.com} \\\And
  Ben Kao \\
  University of Hong Kong \\
  \texttt{kao@cs.hku.hk} \\\And
  {Chao Huang}\thanks{Chao Huang is the corresponding author.} \\
  University of Hong Kong \\
  \texttt{chuang7@hku.hk}
  }

\begin{document}
\maketitle
\begin{abstract}
Graph learning has become essential in various domains, including recommendation systems and social network analysis. Graph Neural Networks (GNNs) have emerged as promising techniques for encoding structural information and improving performance in tasks like link prediction and node classification. However, a key challenge remains: the difficulty of generalizing to unseen graph data with different properties. In this work, we propose a novel graph foundation model, called \model, to address this challenge. Our approach tackles several technical obstacles. Firstly, we enhance data augmentation using a large language model (LLM) to overcome data scarcity in real-world scenarios. Secondly, we introduce a unified graph tokenizer that enables the model to generalize effectively to diverse graph data, even when encountering unseen properties during training. Thirdly, our developed scalable graph transformer captures node-wise dependencies within the global topological context. Extensive experiments validate the effectiveness of our framework. By adapting \model\ to new graph characteristics and comprehending diverse graphs, our approach achieves remarkable zero-shot graph learning performance across various settings.
We release the model implementation at \href{https://github.com/HKUDS/OpenGraph}{https://github.com/HKUDS/OpenGraph}.

\end{abstract}

\input{intro}
\input{model}
\input{solution}
\input{eval}
\input{relate}
\input{conclusion}

\clearpage
\input{limitation}
\bibliography{custom}

\clearpage
\appendix
\input{appendix}

\end{document}

%% file: intro.tex
\section{Introduction}
\label{sec:intro}


Graph learning is a crucial methodology in various fields, such as recommender systems~\cite{he2020lightgcn}, social network analysis~\cite{sankar2021graph}, citation networks~\cite{lv2021we}, and transportation networks~\cite{wang2020traffic}. By utilizing Graph Neural Networks (GNNs) and recursive message passing, we capture the complex structures of graphs effectively. GNNs leverage inter-dependencies among nodes to incorporate high-order connectivities into learned graph representations~\cite{ying2018graph,jin2020graph}.


A primary challenge in current end-to-end graph neural networks is their heavy reliance on scarce and low-quality labeled data~\cite{liu2022graph,jin2021automated}. To overcome this, self-supervised learning (SSL) has emerged as a solution by leveraging augmented self-supervision signals. Contrastive SSL, exemplified by DGI~\cite{velivckovic2018deep} and GraphCL~\cite{you2020graph}, incorporates contrastive objectives as self-supervised alignment loss. Recent advancements like JOAO~\cite{you2021graph} and GCA~\cite{zhu2021graph} automate contrastive learning through adaptive augmentation. By integrating SSL techniques, we enhance graph neural networks with limited labeled data.


Graph pre-training excels at capturing intrinsic graph properties but struggle to effectively generalize to diverse downstream domains, particularly when faced with distribution shifts~\cite{sun2023all,wu2021handling,gui2022good}. For example, in recommender systems, handling previously unseen user interaction graphs in cold-start recommendation scenarios is crucial~\cite{chen2022generative}. Transferring knowledge from pre-trained graph domains to other downstream domains is desirable~\cite{zhang2022few}. However, applying these models to unseen graphs results in significant performance deterioration due to variations in node sets and relation semantics across different scenarios.

Recent research explores prompt-tuning as a task-specific alternative to fine-tuning, bridging the gap between pre-training and downstream objectives~\cite{sun2022gppt,liu2023graphprompt,fang2022universal}. These approaches align the pre-trained model's understanding with specific task requirements. However, practical scenarios involve variations in node sets and feature semantics across diverse downstream graph domains. Further exploration is needed to enhance graph models' generalization and adaptability to real-world graphs.

This work aims to develop a scalable graph model that enables zero-shot learning, effectively making accurate predictions on unseen graphs. Building such a model poses significant challenges.

\begin{itemize}[leftmargin=*]

    \item \textbf{C1}: \textbf{Domain-Specific Data Scarcity}. Data scarcity is a common challenge across downstream domain tasks, driven by factors like privacy concerns. Limited availability of domain-specific user behavior graphs restricts data collection. Therefore, developing label-less learning frameworks within graph models is crucial to effectively understand the context of downstream tasks in the face of data scarcity. \vspace{-0.1in}

    \item \textbf{C2}: \textbf{Node Token Set Shift}. A key challenge in zero-shot graph learning is the shift in node token sets across graphs. This requires the model to reconcile variations in node characteristics. Generating universal graph tokens is crucial to effectively represent and comprehend diverse unseen graphs with different topological contexts. \vspace{-0.25in}

    \item \textbf{C3}: \textbf{Efficient Graph Dependency Modeling}. Nodes in large-scale graphs have complex dependencies. Understanding local and global inter-dependencies among all nodes is crucial for accurate prediction. Efficient node-wise dependency encoding is vital to enhance the performance and scalability of graph models. \vspace{-0.05in}
    
    
\end{itemize}

\noindent \textbf{Present Work}. To overcome the challenges, we introduce a graph model for zero-shot learning that captures universal and transferable structural patterns across multiple domains. To address \textbf{C1}, we propose combining large language models (LLMs) with data augmentation techniques for synthetic graph generation. By generating augmented graphs resembling real-world instances, we enhance the pre-training process of \model\ and gain a deeper understanding of downstream task contexts. This is achieved through the integration of tree-of-prompt regularization with Gibbs sampling.

To address \textbf{C2}, we propose a topology-aware graph tokenizer that generates universal graph tokens from arbitrary graphs. For \textbf{C3}, we develop a scalable graph transformer with efficient self-attention using anchor sampling. Our approach ensures computational efficiency through a two-stage self-attention process and optimizes training by leveraging token sequence sampling, reducing sequence length while preserving crucial context.
We conducted extensive experiments on diverse datasets, demonstrating the remarkable generalization abilities of our model across various settings. 

%% file: model.tex
\section{Preliminaries}
\label{sec:model}


\noindent {\bf Graph Learning}. 
A graph $\graph=(\setv, \sete, \matf)$ consists of a node set $\setv={v_i}$, an edge set $\sete={(v_s, v_t)}$, and node attributes $\matf\in\dmnr^{|\setv|\times f}$. Graph learning aims to produce node representations that encode both structural and attribute information. These embeddings are used for tasks such as link prediction and node classification, involving the prediction of node connections and categories, respectively. The corresponding losses to be minimized are:
\begin{align}
    &\loss_{\text{link}}=\sum_{v_s, v_t} (1-e_{s,t})f(v_s,v_t) - e_{s,t}f(v_s,v_t), \nonumber\\
    &\loss_{\text{node}}=-\sum_{v_s,y_s} \Big({f(v_s, y_s)} ~/ {\sum_{y'_s\neq y_s} f(v_s, y'_s)}\Big)
\end{align}
where $e_{s,t}\in\{0, 1\}$ denotes the link label for nodes $v_s$ and $v_t$, and $y_s\in\setc$ indicate the groundtruth category for node $v_s$. Function $f$ denotes the prediction model with learnable parameters $\param_f$.

\noindent {\bf Zero-shot Graph Learning}.
Current graph models excel in standard tasks but struggle to generalize across diverse domains. Their performance deteriorates when applied to new graphs with varying characteristics, such as node sets and features. To address these limitations, we focus on \emph{zero-shot graph learning}, where a model is trained on a set of graphs and evaluated on different test graphs without shared graph tokens. It aims to assess the model's ability to learn generalized topological structures and node-wise dependencies. Formally, we seek to minimize the error measurement $\epsilon({\graph_t}, f)$, with $\mathop{\arg\min}\nolimits_f$ denoting the optimization.
\begin{align}
    \param_f=&\mathop{\arg\min}\nolimits_{\param_f}\loss(\{\graph_s\}, f), \nonumber\\
    \setv_t \cap \setv_s = &\varnothing,~ \sete_t\cap\sete_s = \varnothing, ~\dmnr^{f_t}\neq\dmnr^{f_s}
\end{align}
This objective is to develop a universal graph modeling architecture $f(\cdot)$. Its parameters $\param_f$ are learned by minimizng graph learning losses $\loss$ on the training graphs $\{\graph_s\}$. Notably, the training graphs and test graphs $\graph_t$ have no common nodes, edges, or node features. This presents a unique challenge for the graph model to handle the significant distribution shift that occurs across different graph domains with entirely distinct datasets.

%% file: solution.tex
\section{Methodology}
\label{sec:solution}

This section presents the design details of the proposed \model\ framework. Figure~\ref{fig:framework} gives an overall illustration for the model. In appendix, we elaborate on how \model\ handles node classifiction (\ref{app:zero_node_class}) and graph features (\ref{app:zero_node_feats}), detailed configurations of our scalable graph transformer (\ref{app:scalable_graph_transformer}), as well as the generation algorithms (\ref{app:generation}).

\begin{figure*}
    \centering
    \includegraphics[width=0.98\textwidth]{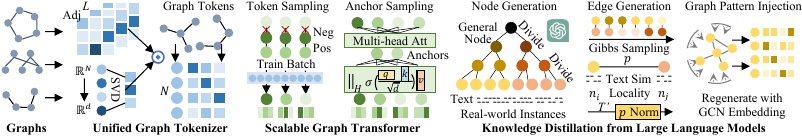}
    \vspace{-0.1in}
    \caption{Overall model architecture of the \model\ framework.}
    \vspace{-0.15in}
    \label{fig:framework}
\end{figure*}

\subsection{Unified Graph Tokenizer}
To handle diverse graphs with varying nodes and features, our goal is to develop a graph tokenizer that transforms input graphs into unified token sequences: $\graph\rightarrow\{\vece_i\}$. Each token represents a node accompanied by a semantic vector $\vece_i$. By utilizing a shared representation space and a flexible sequence structure, we aim to standardize distributions across graphs.
Specifically, our tokenizer uses the smoothed adjacency matrix $\tilde{\mata}$, and a topology-aware projection function $\phi:\dmnr^{|\setv|} \rightarrow \dmnr^d$.

\subsubsection{\bf Smoothed High-Order Adjacency}
We start with the original adjacency matrix $\mata\in\dmnr^{|\setv|\times|\setv|}$ built from edges $\sete$. The smoothing procedure for the adjacency matrix is as follows:
\begin{align}
    \tilde{\mata} = \bar{\mata}^1 + \bar{\mata}^2 + \cdots \bar{\mata}^L,~~~~~
    \bar{\mata} = \matd^{-\frac{1}{2}} \mata \matd^{-\frac{1}{2}}
\end{align}
For numerical stability, we use Laplacian normalization $\bar{\mata}$ with the diagonal degree matrix $\matd$ of adjacency $\mata$. To capture high-order connectivity and sparse node-wise relations, \model\ combines $\bar{\mata}$ at different orders. This provides us with topology information for further processing, with $L$ representing the maximum power order considered.

\subsubsection{\bf Topology-aware Projection with Arbitary Graphs}
To handle the varying dimensionality $|\setv|\times|\setv|$ of adjacency $\tilde{\mata}$, \model\ applies a projection function $\phi: \mathbb{R}^{|\setv|} \rightarrow \mathbb{R}^d$ to transform the adjacency into sequence data. A large hidden dimensionality $d$ is used to minimize information loss. Previous research has shown that even random projections with large dimensions can achieve satisfactory performance~\cite{zheng2022instant}. To preserve topology information, we employ fast singular value decomposition (SVD) as the projection $\phi$. SVD is known for its efficiency and effectiveness in adjacency compression~\cite{jamali2010matrix}. Our empirical analysis demonstrates that two iterations of fast SVD effectively preserve topology information with minimal computational overhead. The graph tokenizer performs the following operations to calculate the resulting token sequence:
\begin{align}
    \vece_v = \phi(\tilde{\mata}_{v,:}) = \tilde{\mata}_{v,:} \cdot \text{LN}((\matu\sqrt{\Lambda} ~ \| ~ \matv\sqrt{\Lambda}))
\end{align}
where $\matu, \matv \in\dmnr^{|\setv|\times d}$ and $\Lambda\in\dmnr^{d\times d}$ are obtained from SVD. The concatenation operator $\|$ combines them in the hidden dimension. Layer normalization function $\text{LN}(\cdot)$ reduces numerical variance across datasets. The resulting $\vece_v\in\dmnr^d$ incorporates topology information from $\tilde{\mata}$ and the topology-aware projection $\phi$. This information strengthens subsequent learnable neural networks.



\subsection{Scalable Graph Transformer}


With the universal topology-aware graph tokens, the subsequent task is to empower our graph model to grasp the complex node-wise dependencies within the global context. Inspired by the success of transformer architectures in modeling complex relationships between instances, \model\ utilizes a graph transformer as the backbone. To ensure scalability and effectiveness for large-scale graphs, we introduce the following techniques.

\noindent\textbf{Token Sequence Sampling}. For efficiency, we train the graph transformer using sampled token sequences from the current training batch, which contains centric nodes $v_{c_b}$, positive nodes $v_{p_b}$, and negative nodes $v_{n_b}$. The input is as follows:
\begin{align}
    (\vece_{c_1}\cdots \vece_{c_B}) ~\|~ (\vece_{p_1}  \cdots \vece_{p_B}) ~\|~ (\vece_{n_1} \cdots \vece_{n_B})
\end{align}
This approach significantly reduces the sequence length from $|\setv|$ to $3\times B$, enabling efficient training for large-scale graphs. Despite using a sub-sequence, the topology-aware embeddings contain local structural information for each node and reflect the overall graph structure. Additionally, this sampling technique emphasizes the current training batch, leading to further training improvements.


\noindent\textbf{Efficient Self-Attention with Anchors}.
To accelerate the self-attention part of \model\ with quadratic complexity, we introduce a step of sampling anchor nodes $v_{a_s}$ for $s\in S$, where $S<3B$. This splits the self-attention process into two stages: propagating messages from all nodes to the anchor nodes and then propagating the anchor embeddings to all nodes. This decomposition reduces the complexity from $\complexity(B^2\times d)$ to $\complexity(B\times S)$, ensuring scalability for large-scale graphs.

\subsection{Knowledge Distillation from LLM}
\label{sec:gen_data}
Obtaining diverse graph datasets for different domains can be challenging due to factors like privacy issues that restrict access to essential data~\cite{zheleva2007preserving}. Inspired by the remarkable knowledge and understanding demonstrated by large language models (LLMs), we leverage their power to enhance the generation of diverse graph-structured data. To improve the efficacy of our LLM-augmented graph data for pre-training our model, we have developed an augmentation mechanism. This mechanism enables the LLM-augmented graph data to closely approximate real-world graph characteristics, enhancing the relevance and usefulness of the augmented data.


\subsubsection{\bf LLM-based Node Generation}
Our first step is to create a node set tailored to the application scenario, characterized by text-based profiles that generate subsequent edges. However, dealing with real-world scenarios poses challenges due to the large scale of the node set. For example, e-commerce platforms may have billions of products, making it challenging for the LLM to efficiently generate a large number of nodes.

To address this challenge, we adopt an iterative strategy of dividing general nodes into sub-categories with finer semantic granularity. For instance, in the case of generating product nodes, we prompt the LLM with a query like "List sub-categories of \emph{products} on platforms like Amazon." The LLM provides a list of sub-categories such as "clothing" and "electronics." We repeat this iterative division process, refining each sub-category further, until we obtain nodes that resemble real-world instances, such as "women's clothing," "sweaters," "hooded sweaters," and "white hooded sweaters." Appendix~\ref{app:prompt} presents details on our prompt template and generation examples.


\noindent\textbf{Tree-of-Prompt Algorithm}. The process of dividing nodes into sub-categories and generating fine-grained entities follows a tree structure. The initial general node (e.g., "products," "deep learning papers") serves as the root, and fine-grained entities act as leaf nodes. We employ a tree-of-prompt strategy to traverse and generate these nodes. For further details, please see Appendix~\ref{app:dfs}.

\vspace{-0.05in}
\subsubsection{\bf Edge Sampling using Node Profiles}
To generate edges, we use the Gibbs sampling algorithm~\cite{gelfand2000gibbs} with the generated node set $\setv$. The algorithm starts with a random sample.
For instance, in a paper-wise citation network, the initial sample is a node pair $(v_{s_0}, v_{t_0})$. In a person-entity relation scenario like an author-paper network, the initial sample is a binary vector $\veca^0\in\{0,1\}^{|\setv|}$. Each element $a_i\in\veca^0$ indicates whether there is an interaction between the sampled person and the $i$-th node $v_i$. In the case of person-entity relations, the Gibbs algorithm for edge sampling is described in Appendix~\ref{app:gibbs}. The key is estimating the probability $p(\mathbf{a}^t \oplus v_{t'}|\mathbf{a}^t)$, with $\oplus$ representing setting the $t'$-th dimension of $\veca^t$ to 1.

\noindent\textbf{Node-wise Connection Probability Estimation}.
To estimate the probability $p(\mathbf{a}^t \oplus v_{t'}|\mathbf{a}^t)$ of connecting two nodes in our generated graph, we leverage the reasoning capabilities of the LLM. However, directly prompting the LLM for predictions on each edge can be computationally expensive, with $\mathcal{O}(|\mathcal{V}|\times |\mathcal{V}|)$ prompts required. To ensure efficiency, we adopt an alternative approach. We prompt the LLM to generate hidden representations $\mathbf{h}_i$ for each node $v_i$. Then, we calculate the probability for each edge with dot-product as:
\begin{align}
    p(\veca^t\oplus v_{t'}|\veca^t) = \sum\nolimits_{v_{i}} a^t_i (\vech_i / \|\veca^t\|_0)^\top \cdot \vech_{t'}
\end{align}
By utilizing the text embeddings $\vech_i$ and $\vech_{t'}$ provided by the LLM, we can effectively capture the semantic relations between the respective nodes.

\noindent\textbf{Dynamic Probability Normalization}. 
To ensure that the calculated probability scores fall within a reasonable range like $[0, 1]$, our generation algorithm incorporates a dynamic probability normalization approach.
It maintains a record of the most recent $T'$ estimation values, denoted as $\setp = \{p(\veca^t\oplus v_{t'}|\veca^t)~|~t=-1,\cdots,-T'\}$. By calculating the mean ($\mu$) and standard deviation ($\sigma$) of these values, we gain insight into their distribution. New estimations are adjusted using $\mu \pm 2\sigma$ as the bounds, resulting in $\bar{p} = (p - \mu) / (4\sigma)$.

\noindent\textbf{Node Locality Incorporation}.
To address the limitation of the previous edge sampling algorithm based on semantic similarity, we introduce the concept of locality. Each node is assigned a random locality index, and we consider the difference in locality using an exponential decay function. This results in an adjusted probability calculated as $\hat{p} = \bar{p} \cdot \alpha^{|n_i - n_j|}$, where $0 < \alpha < 1$. By incorporating locality, we account for the observed patterns in real-world graphs and prevent excessive connections among semantically-related nodes.

\subsubsection{\bf Graph Topological Pattern Injection}
To enhance the incorporation of topological information in the graph generation process, we refine the node embeddings after the initial graph generation. By training a Graph Convolutional Network (GCN) on the graph $\graph$, we obtain new node embeddings that capture the underlying topology patterns. This aligns the node embeddings derived from the graph with the textual embeddings of the entities and avoids distribution shifts between graph and textual spaces. The final graph is constructed using our edge generation algorithm, which operates on these enhanced node representations.



%% file: eval.tex
\section{Evaluation}
\label{sec:eval}

\subsection{Experimental Settings}
\noindent\textbf{Datasets}. We evaluate \model\ on two graph learning tasks: link prediction and node classification, using totally 8 real-world datasets. Appendix~\ref{app:dataset} provides detailed descriptions.

\noindent\textbf{Evaluation Protocol}. Following previous works~\cite{he2020lightgcn, kipf2016semi}, we adopt the original train-test data split for the experimental datasets. We pre-train our \model\ on generated datasets and conducts zero-shot prediction for the evaluation datasets made of real graph data. As most baselines struggle with cross-dataset transferring, we evaluate them in two few-shot training settings. Please refer to Appendix~\ref{app:evaluation_protocol} for more details about our cross-dataset zero-shot setting, few-shot settings, and evaluation metrics.

\noindent\textbf{Implementation Details}. We provide detailed information about the implementation of \model\ and the baseline methods, as well as the graph generation process, in Appendix~\ref{app:implementation}.

\begin{table*}[t]
    \centering
    \caption{Performance of \model\ (zero-shot) and baseline methods (one-shot, five-shot) on link prediction (measured by \textit{Recall@N} for $N=20, 40$) and node classification (measured by \textit{Accuracy} and \textit{Macro F1 Score}).}
    \label{tab:overall_performance}
    \small
    \setlength{\tabcolsep}{0.7mm}
    \vspace{-0.1in}
    \begin{tabular}{c|c|c|c|c|c|c|c|c|c|c|c||c|c|c|c|c|c}
        \hline
        \multicolumn{2}{c|}{Dataset} & \multicolumn{2}{c|}{ogbl-ddi} & \multicolumn{2}{c|}{ogbl-collab} & \multicolumn{2}{c|}{ML-1M} & \multicolumn{2}{c|}{ML-10M} & \multicolumn{2}{c||}{Amazon-book} & \multicolumn{2}{c|}{Cora} & \multicolumn{2}{c|}{Citeseer} & \multicolumn{2}{c}{Pubmed}\\
        \hline
        Model & shot & R-20 & R-40 & R-20 & R-40& R-20 & R-40& R-20 & R-40& R-20 & R-40 & Acc & F1 & Acc & F1 & Acc & F1\\
        \hline
        \hline
        \multirow{2}{*}{MF} & 1 & .0087 & .0161 & .0261 & .0349 & .0331 & .0604 & .1396 & .1956 & .0034 & .0043 & .1710 & .1563 & .1740 & 	.1727 & .3470 &  .3346\\
        & 5 & .0536 & .0884 & .0412 & .0609 & .0987 & .1584 & .2060 & .2989 & .0196 & .0284 & .1500 & .1422 & .1520 & .1484 & .3540 & .3435\\
        \hline
        \multirow{2}{*}{MLP} & 1 & .0195 & .0336 & .0112 & .0185 & .0548 & .1019 & .1492 & .2048 & .0017 & .0028 & .2300 & .1100 & .2590 & .1993 & .4430 & .3114\\
        & 5 & .0621 & .1038 & .0115 & .0185 & .0851 & .1470 & .2362 & .2563 & .0092 & .0152 & .3930 & .3367 & .3690 & .3032 & .5240 & .4767\\
        \hline
        \multirow{2}{*}{GCN} & 1 & .0279 & .0459 & .0206 & .0321 & .0432 & .0849 & .1760 & .2086 & .0096 & .0160 & .3180 & .1643 & .3200 & .2096 & .4270 & .3296\\
        & 5 & .0705 & .1312 & .0366 & .0513 & .1054 & .1656 & .2127 & .2324 & .0251 & .0408 & .5470 & .5008 & .4910 & .4190 & .509 & .4455\\
        \hline
        \multirow{2}{*}{GAT} & 1 & .0580 & .1061 & .0258 & .0372 & .0245 & .0520 & .1615 & .2476 & .0047 & .0079 & .2420 & .1687 & .2810 & .2025 & .4720	& .3657\\
        & 5 & .0711 & .1309 & .0340 & .0505 & .1506 & .2267 & .2002 & .2883 & .0228 & .0392 & .585 & .5438 & .4940 & .4441 & .5780 & .5582\\
        \hline
        \multirow{2}{*}{GIN} & 1 & .0530 & .1004 & .0163 & .0247 & .0466 & .0884 & .1541 & .2388 & .0069 & .0114 & .3190 & .1753 & .2820 & .1705 & .4410 & .3064\\
        & 5 & .0735 & .1441 & .0311 & .0458 & .1458 & .2344 & .1926 & .2829 & .0252 & .0418 & .5400 & .4941 & .521 & .4696 & .5070 & .4547\\
        \hline
        \multirow{2}{*}{DGI} & 1 & .0315 & .0617 & .0255 & .0385 & .0486 & .0863 & .1868 & .2716 & .0081 & .0142 & .3150 & .1782 & .2840 & .1791 & .4290 & .3163\\
        & 5 & .0821 & .1426 & .0345 & .0502 & .1687 & .2573 & .2303 & .3063 & .0300 & .0492 & .4880 & .4606 & .4450 & .4062 & .4890 & .4509\\
        \hline
        \multirow{2}{*}{GPF} & 1 & .0503 & .0856 & .0027 & .0048 & .1099 & .1702 & .1599 & .2326 & .0072 & .0128 & .3080 & .1952 & .3110 & .1984 & .4220 & .2670\\
        & 5 & .0839 & .1460 & .0027 & .0047 & .0817 & .1392 & .2014 & .2994 & .0179 & .0310 & .5550 & .5233 & .4690 & .4223 & .5150 & .4934\\
        \hline
        \multirow{2}{*}{GPrompt} & 1 & .0541 & .1102 & .0138 & .0207 & .0797 & .1310 & .1362 & .2073 & .0074 & .0120 & .3540 & .1596 & .2800 & .1519 & .4710 & .3705\\
        & 5 & .0769 & .1396 & .0157 & .0231 & .1340 & .2166 & .2157 & .3147 & .0287 & .0464 & .5510 & .5098 & .5570 & .5211 & .5130 & .4520\\
        \hline
        \multirow{2}{*}{GraphCL} & 1 & .0603 & .1112 & .0265 & .0398 & .0390 & .0799 & .1655 & .2529 & .0047 & .0077 & .2430 & .1548 & .2980 & .1630 & .4070 & .4130\\
        & 5 & .0740 & .1368 & .0311 & .0456 & .1416 & .2138 & .2019 & .3075 & .0270 & .0440 & .5610 & .5330 & .4300 & .3683 & .5230 & .5024\\
        \hline
        \hline
        Ours & 0 & \textbf{.0921} & \textbf{.1746} & \textbf{.0421} & \textbf{.0639} & \textbf{.1911} & \textbf{.2978} & \textbf{.2370} & \textbf{.3265} & \textbf{.0485} & \textbf{.0748} & \textbf{.7504} & \textbf{.7426} & \textbf{.7221} & \textbf{.6801} & \textbf{.6869} & \textbf{.6537}\\
        \hline
    \end{tabular}
    \vspace{-0.1in}
\end{table*}

\noindent\textbf{Baselines}. Our empirical evaluation utilizes the following 9 state-of-the-art baseline methods from 4 different research lines. Detailed descriptions for the baselines can be found in Appendix~\ref{app:baselines}.

\subsection{Overall Performance Comparison (RQ1)}
Comparing the zero-shot performance of \model\ with the few-shot performance of baselines in link prediction and node classification (Table~\ref{tab:overall_performance}), we have the following observations:

\noindent \textbf{Predominant Performance of \model}. 
Our model outperforms baselines on all 8 datasets in different categories without using any overlapping data between pre-training and downstream tasks, showcasing its remarkable ability to generalize. This advantage can be attributed to three key factors: i) the unified graph tokenizer, bridging the gap between pre-training and target datasets; ii) the scalable graph transformer, capturing important structural features and learning relations effectively; and iii) effective pre-training with LLM-generated graph data, equipping our model with versatile forecasting abilities. Overall, these design choices contribute to our model's outstanding generalization capabilities across diverse datasets.

\noindent \textbf{Limitations of Existing Pretraining Methods}. 
Pre-training methods like GraphCL and DGI do not consistently outperform foundational models (e.g., GIN and GCN) trained on few-shot data, suggesting that they may hinder performance when applied to different datasets. This is due to a significant shift in data distribution between pre-training and target datasets, leading to overfitting and impairing adaptability to new graph structures in downstream tasks. Prompt tuning methods like GraphPrompt and GPF sometimes degrade more severely compared to full-model fine-tuning approaches, indicating a higher susceptibility to overfitting to distinctive patterns within the training set.

\noindent \textbf{Tackling Node Classification by Structure Learning}. Our \model\ shows significant improvement in node classification tasks, highlighting the effectiveness of using our pre-trained link predictor to identify connections between ordinary nodes and special nodes representing classes. This advantage relies on \model's strong ability to generalize, transferring knowledge across datasets and tasks. The credit for this versatility goes to the universal applicability and flexibility of our graph tokenization and encoding.


\subsection{Investigation on Graph Tokenizer (RQ2)}
In this section, we study the effectiveness of our graph tokenizater by evaluating the impact of the smoothed adjacency matrix and comparing our projection method with alternative compression methods. The results are summarized in Figure~\ref{fig:graph_tokenizer}.

\noindent \textbf{Impact of adjacency smoothing}: We examine the effect of graph smoothing on model performance by testing different levels of smoothing for the input adjacency matrix. The results are depicted in Figure~\ref{fig:ddi_adj_order} and~\ref{fig:ml1m_adj_order}. Here, the use of 0 adjacency smoothing implies the input of an identity matrix for the graph tokenizer. This approach significantly damages the topological information for the graph tokenizer, resulting in poor performance. This outcome underscores the importance of considering the adjacency matrix within our unified graph tokenizer.
For the non-zero graph smoothing orders, $L=2$ produces the best performance for the Movielens-1M dataset. $L=3$ and $1$ yield the best performance for the OGBL-ddi data under the top-20 and top-40 settings, respectively. This suggests the benefits of exploring high-order graph smoothing in the graph tokenizer of our \model.

\noindent\textbf{Superiority of topology-aware projection}. To assess the effectiveness of our topology-aware projection based on SVD, we compare it to three alternative projection methods (see Appendix~\ref{app:proj_ablation} for details). The results are presented in Figure~\ref{fig:ddi_proj} and~\ref{fig:ml1m_proj}. We make the following observations:\vspace{-0.08in}
\begin{itemize}[leftmargin=*]
    \item \textbf{One-hot encoding}. This approach learns id-corresponding embeddings across datasets. It performs poorly in the zero-shot evaluation, highlighting the difficulty of transferring dataset-specific parameters like node embeddings to unseen datasets that lack overlapping node tokens.\vspace{-0.12in}
    \item \textbf{Degree embeddings}. This method learns degree-specific embeddings. It performs significantly worse than our projection scheme. This is because there is a substantial semantic gap for the same degree number across different graphs. Moreover, it oversimplifies topology features by considering only the number of direct links, limiting its ability to capture nuanced structural patterns and adversely affecting graph projection.\vspace{-0.12in}
    \item \textbf{Random projection}. It randomly assigns unlearnable embedding vectors to nodes. It outperforms the other two variants, but its performance is still inferior to our method due to the low representation efficiency of its uniform distribution.

\end{itemize}

\subsection{Influence of Pre-training Datasets (RQ3)}
To evaluate the effectiveness of our knowledge distillation from the LLM, we compare the performance of \model\ networks pre-trained with different datasets. We use three ablated versions of our graph generation algorithm, namely -Norm, -Loc, and -Topo. Additionally, we incorporate two real datasets, Yelp2018 and Gowalla, for pre-training, which are unrelated to the test datasets. The ML-10M dataset, related to ML-1M and itself, is also included. The evaluation results are summarized in Table~\ref{tab:pretrain_dataset}. We draw the following conclusions:

\begin{figure}[t]
    \centering

    \subfigure[Recall@20]{
        \includegraphics[width=0.24\columnwidth]{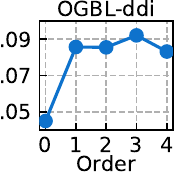}
        \hspace{-0.08in}
        \includegraphics[width=0.24\columnwidth]{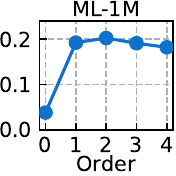}
        \label{fig:ddi_adj_order}
    }
    \hspace{-0.14in}
    \subfigure[Recall@40]{
        \includegraphics[width=0.24\columnwidth]{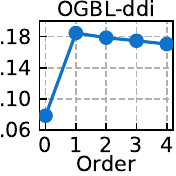}
        \hspace{-0.08in}
        \includegraphics[width=0.24\columnwidth]{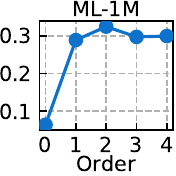}
        \label{fig:ml1m_adj_order}
    }

    \includegraphics[width=0.7\columnwidth]{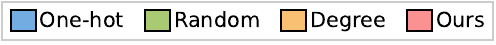}
    \vspace{-0.08in}
    
    \subfigure[Recall@20]{
        \includegraphics[width=0.24\columnwidth]{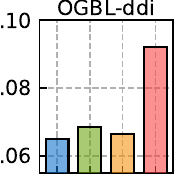}
        \hspace{-0.08in}
        \includegraphics[width=0.24\columnwidth]{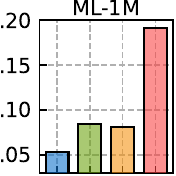}
        \label{fig:ddi_proj}
    }
    \hspace{-0.14in}
    \subfigure[Recall@40]{
        \includegraphics[width=0.24\columnwidth]{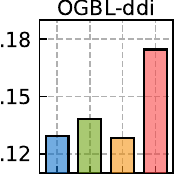}
        \hspace{-0.08in}
        \includegraphics[width=0.24\columnwidth]{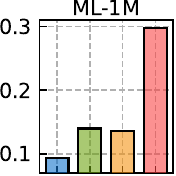}
        \label{fig:ml1m_proj}
    }
    \vspace{-0.12in}
    \caption{Influence of graph tokenizer configurations.}
    \label{fig:graph_tokenizer}
    \vspace{-0.1in}
\end{figure}

\noindent \textbf{Superiority of our generated data}. Our generated dataset (Gen) achieves the best performance on all test datasets except for ML-1M and ML-10M. Notably, ML-10M, which is closely related to these two datasets, achieves the best performance in these cases. This finding highlights the superior generalization ability of our generated datasets, which equips the \model\ model with the capability of universal topological structure learning.

\noindent \textbf{Impact of individual generation techniques}. We conduct ablation study by removing the dynamic probability normalization (-Norm), locality incorporation (-Loc), and graph topological pattern injection (-Topo) modules. The removal of these modules from the generation algorithm leads to a significant drop in performance for the downstream prediction. This demonstrates the importance of these key modules for our generated graphs.

\noindent \textbf{Real-world data can be misleading}. Using real-world datasets (Yelp and Gowalla) does not yield transferable graph learning capabilities that provide an advantage over our Gen data. This highlights the limitation of relying solely on insufficient or biased real-world data for cross-data pre-training.

\noindent \textbf{Pre-training on related datasets is useful}. Models pre-trained on the ML-10M dataset exhibit superior performance when testing on ML-1M and ML-10M, which are directly related datasets. This indicates that dataset-wise similarity can aid in the cross-dataset knowledge transferring.

\begin{table}[]
    \centering
    \small
    \setlength{\tabcolsep}{0.4mm}
    \caption{Impact of using different pre-training datasets.}
    \label{tab:pretrain_data}
    \vspace{-0.12in}
    \label{tab:pretrain_dataset}
    \begin{tabular}{c|ccccccc}
        \hline
        Test & \multicolumn{7}{c}{Pre-training Dataset}\\
        \cline{2-8}
        Data & -Norm & -Loc & -Topo & Yelp & Gowalla & ML10M & Gen\\
        \hline
        \hline
        ogbl-ddi & .0737 & \underline{.0893} & .0656 & .0588 & .0770 & .0692 & \textbf{.0921}\\
        ML1M & .0572 & .1680 & .0850 & .0599 & .0485 & \textbf{.2030} & \underline{.1911}\\
        ML10M & .0982 & .1636 & .1017 & .1629 & .0910 & \textbf{.2698} & \underline{.2370}\\
        Cora & .4985 & .4864 & .4342 & .3715 & \underline{.5943} & .2780 & \textbf{.7504}\\
        Citeseer & .3944 & .3691 & \underline{.5743} & .2651 & .4300 & .2003 & \textbf{.7221}\\
        Pubmed & .4501 & .5015 & .4876 & .3317 & \underline{.5148} & .3652 & \textbf{.6869}\\
        \hline
    \end{tabular}
    \vspace{-0.1in}
\end{table}

\subsection{Impact of Sampling in Transformer (RQ4)}
We examine the influence of token sequence sampling and anchor sampling in our scalable graph transformer architecture. The metrics for efficiency and performance are summarized in Table~\ref{tab:efficiency}. Our evaluation focuses on the GPU memory costs and the running time during both the training and testing processes. 
Additionally, we assess the model performance after end-to-end training. The following observations are made for the ablated versions.\vspace{-0.08in}
\begin{itemize}[leftmargin=*]
    \item \textbf{-S-A}: It eliminates both token sequence sampling and anchor sampling, leading to out-of-memory (OOM) errors when applied to the larger ML-10M data. When evaluated on OGBL-ddi, it exhibits the lowest memory and time efficiency, while its performance is inferior to \model. \vspace{-0.12in}
    \item \textbf{-Anc}: This version incorporates only sequence sampling. It significantly reduces memory costs during the training phase. Additionally, by focusing on the current training context, it achieves the best performance on OGBL-ddi.\vspace{-0.12in}
    \item \textbf{-Seq}: This model removes token sequence sampling from \model, resulting in a significant decrease in training efficiency. However, the anchor sampling strategy in this version greatly reduces computational costs during the test phase. Despite the computational benefits, the anchor sampling strategy leads to a drop in performance compared to the full-version \model.
\end{itemize}

\begin{table}[]
    \centering
    \small
    \setlength{\tabcolsep}{0.8mm}
    \caption{Impact of sampling strategies on the efficiency and performance in the scalable graph transformer.}
    \vspace{-0.12in}
    \label{tab:efficiency}
    \begin{tabular}{ccccccc}
        \toprule
        \multirow{2}{*}{OGBL-ddi} & \multicolumn{2}{c}{Memory} & & \multicolumn{2}{c}{Time} & \multirow{2}{*}{R@20}\\
        \cline{2-3}\cline{5-6}
        & Train & Test & & Train & Test & \\
        \midrule
        -S-A & 5420MiB & 1456MiB & &22.72s & 13.88s & 0.0966\\
        -Anc & 3360MiB & 1456MiB & &18.19s & 13.73s & 0.1107\\
        -Seq & 2456MiB & 1202MiB & &16.45s & 12.09s & 0.0930\\
        Ours & 2358MiB & 1202MiB & &15.45s & 12.09s & 0.1006\\
        \bottomrule
        \multirow{2}{*}{ML-10M} & \multicolumn{2}{c}{Memory} & & \multicolumn{2}{c}{Time} & \multirow{2}{*}{R@20}\\
        \cline{2-3}\cline{5-6}
        & Train & Test & & Train & Test & \\
        \midrule
        -S-A & OOM & OOM & & -- & -- & --\\
        -Anc & 4996MiB & OOM & & 73.15s & -- & --\\
        -Seq & 23140MiB & 4550MiB & & 158.60s & 84.78s & 0.2772\\
        Ours & 4470MiB & 4550MiB & & 68.79s & 54.17s & 0.2816\\
        \bottomrule
    \end{tabular}
    \vspace{-0.15in}
\end{table}

\subsection{Impact of Model Scale (RQ5)}
\begin{figure}[t]
    \centering
    \subfigure[Performance and time change on the OGBL-ddi dataset.]{
        \includegraphics[width=0.51\columnwidth]{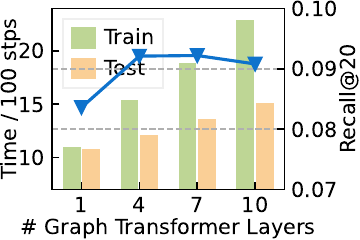}
        \hspace{-0.07in}
        \includegraphics[width=0.51\columnwidth]{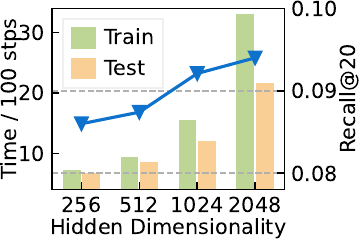}
    }
    
    \subfigure[Performance and time change on the ML-1M dataset.]{
        \includegraphics[width=0.51\columnwidth]{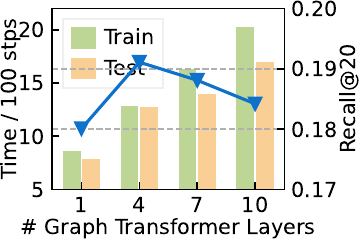}
        \hspace{-0.07in}
        \includegraphics[width=0.51\columnwidth]{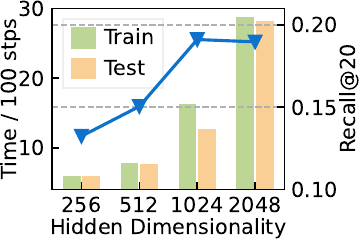}
    }
    \vspace{-0.15in}
    \caption{The impact of model scale on downstream performance and training/testing time (seconds).}
    \label{fig:model_scale}
    \vspace{-0.2in}
\end{figure}

This section examines the influence of model scale within our \model\ framework. Specifically, we modify two crucial hyperparameters that significantly impact the scale of learnable parameters: the number of graph transformer layers $L'$, and the hidden dimensionality $d$. We assess the model's performance on the link prediction task by measuring Recall@20. Additionally, we evaluate the computational time for 100 training steps and 100 test steps. The results are illustrated in Figure~\ref{fig:model_scale}. We summarize the key findings as follows:

\noindent \textbf{Number of graph transformer layers}. It can be observed that both training and testing time of \model\ exhibit a linear increase as the number of graph transformer layers grows. However, the expansion in model size does not consistently lead to performance improvement. The lack of performance enhancement can be attributed to the overfitting effect and the increased training complexity associated with deep transformer networks.

\noindent \textbf{Hidden dimensionality}. In contrast to the number of transformer layers, the hidden dimensionality leads to quadratic growth in computational time. This reflects the rapid increase in model capacity required to accommodate more complex structural data. Consequently, we observe significant improvements in model performance, surpassing the performance curve for the graph transformer layers. Despite the increased model capacity, the growth in hidden dimensionality facilitates the $\dmnr^N\rightarrow\dmnr^d$ projection, reducing information loss and enhancing the quality of the graph token sequence for the subsequent graph transformer layers.

\subsection{End-to-End Training Performance (RQ6)}
To assess the modeling capabilities of our \model\ framework, we perform a performance comparison between \model\ and the baselines trained on the same few-shot datasets. Due to space limitations, we present the detailed results and analysis in Appendix~\ref{app:end_to_end_comparison}. The results demonstrate strong graph learning capabilities of our \model, even in the supervised learning settting.

%% file: relate.tex
\section{Related Work}
\label{sec:relate}

\noindent \textbf{Graph Neural Networks} have gained attention for their ability to model complex relations in graphs~\cite{wu2020comprehensive,chen2020simple}. GNNs use message passing to propagate information from neighboring nodes~\cite{jin2021universal,yuan2020xgnn}. Representative methods include Graph Convolutional Networks (GCNs)~\cite{gao2018large,zhang2021lorentzian} and Graph Attention Networks (GATs)~\cite{zhang2022graph,liao2019efficient}. \model\ is inspired by the Graph Transformer~\cite{yun2019graph,hu2020heterogeneous}, known for capturing global dependencies in graphs.


\vspace{0.03in}
\noindent\textbf{Self-Supervised Graph Learning} aims to address the limited labeled data issue in graph tasks. These methods leverage graph structure and patterns for data-efficient training~\cite{wu2021selftkde,lee2022augmentation,xia2023automated,xiao2022decoupled,yang2023graph}. Graph contrastive learning frameworks, such as GraphCL~\cite{you2020graph} and SGL~\cite{wu2021self}, create meaningful representations by contrasting positive and negative samples using stochastic data augmenters. Adaptive augmentation schemes like JOAO~\cite{you2021graph} and GCA~\cite{zhu2021graph} have been proposed. DGCL \cite{li2021disentangled} and UMGRL \cite{mo2023disentangled} address disentangling factors in contrastive learning. However, these solutions struggle with generalization. \model\ enhances graph model generalization across different tasks.


\vspace{0.03in}
\noindent \textbf{LLM-based Graph Analysis}. Recent advancements in LLMs have prompted interest in utilizing them for enhanced graph comprehension and analysis~\cite{ren2023representation}. GraphLLM~\cite{chai2023graphllm} and GraphQA~\cite{fatemi2023talk} transform graphs into natural language descriptions, enabling improved interpretation and reasoning with LLMs. Techniques like instruction tuning in GraphEdit~\cite{guo2024graphedit} and GraphGPT~\cite{tang2023graphgpt} incorporate rich textual information from text-attributed graphs for fine-tuning LLMs. However, in certain domains like user behavior graphs and neuronal graphs, obtaining high-quality textual features associated with graph nodes can be challenging. Therefore, there is a need for a graph model that can capture universal structural patterns from graphs, even in the absence of textual data.

%% file: conclusion.tex
\section{Conclusion}
\label{sec:conclusion}

This research aims to develop an adaptable framework for capturing complex topological patterns in diverse graph structures. Our model demonstrates exceptional generalization capabilities in zero-shot graph learning tasks across various applications. We utilize a scalable graph transformer architecture and LLM-enhanced data augmentation for efficiency and robustness. Extensive experiments on benchmark datasets validate our model's performance. Future plans include incorporating counterfactual learning to discover noisy connections and influential structures while learning universal and transferable structural patterns in diverse graphs.

%% file: limitation.tex
\section{Limitations}
Our study serves as an initial exploration of graph foundation models, focusing on distilling the generalization capabilities from LLMs without relying on textual features. However, it is crucial to acknowledge and address the limitations that require further attention in future studies.

\emph{Firstly}, it is important to note that \model\ currently does not include modeling for heterogeneous relations and node types. This limitation may impact its performance and generalization capabilities, particularly when dealing with graph data that exhibits strong heterogeneity, such as knowledge graphs. Future research should prioritize the incorporation of heterogeneous representation learning modules to enable a wider range of applications.

\emph{Secondly}, graphs are highly versatile data structures that can be applied to a wide range of domains. While we have evaluated the performance of \model\ on 8 datasets spanning different domains, it is crucial to further strengthen the experimental validation by testing it on additional datasets from even more diverse application domains. This will provide a more comprehensive understanding of its effectiveness and applicability.

\emph{Lastly}, while \model\ shows promising generalization capability, its explainability remains unexplored. This not only hinders its applicability due to its black box nature, but also hinders us from gaining a deeper understanding of the underlying principles that drive its strong generalization capabilities. Future research should prioritize investigating techniques and methodologies to enhance the explainability of \model, allowing researchers and practitioners to gain insights into the internal workings of the model and to ensure its reliable and transparent deployment in real-world applications.

In summary, future studies should focus on incorporating heterogeneous representation learning, expanding the range of tested datasets, and enhancing the explainability of \model\ for broader applicability and deeper insights.

%% file: appendix.tex
\section{Appendix}
\label{tab:appendix}
The appendix offers additional details about our \model\ framework. It covers the neural model component, the generation algorithm, as well as the experimental settings and supplementary results.

\subsection{Methodology}
\subsubsection{Zero-Shot Node Classification}
\label{app:zero_node_class}
Node classification tasks face challenges when transferring trained classification capabilities from one graph to another due to the heterogeneity of node classes across datasets. To address this challenge, our \model\ transforms the graph-specific classification task into a unified link prediction task. This task involves predicting links between regular nodes and special nodes representing different classes. By leveraging the generalized topology extraction capability and learning from observed class-node relations, our \model\ enables zero-shot node classification.

\subsubsection{Handling Node Features}
\label{app:zero_node_feats}
In attributed graphs, node attributes can vary across different graphs, including textual, numerical, and categorical features. To address the semantic differences in node features, our \model\ transforms these attributes into a unified graph structure format. This format can be easily tokenized by our unified graph tokenizer and comprehended by our trained graph transformer. In our approach, we sample node pairs with the highest similarity scores as augmented edges. The similarity scores $s_{i,j}$ between a node pair $(v_i, v_j)$ are calculated as $s_{i,j}=\textbf{f}_i^\top \textbf{f}_j$. Using these similarity scores, we select the top $B\times K$ edges for each batch, where each batch contains $B\times |\mathcal{V}|$ candidate edges. This strategy is applied to all experimental datasets that have node features.

\subsubsection{Details of Scalable Graph Transformer}
\label{app:scalable_graph_transformer}
In the efficient self-attention with anchors, we determine the number of anchors $S$ by $S=d/H<B$, to ensure that the self-attention module incurs similar memory costs as other fully-connected components. Here, $H$ represents the number of attention head. More specifically, the self-attention process for each head can be summarized as follows:
\begin{align}
    \vece_t^{(3)} &= \sum_{v_a} \alpha_{t, a} \matw^{(v)} \vece_a^{(2)},~
    \vece_a^{(2)} = \sum_{v_t} \alpha_{a, t} \matw^{(v)} \vece_t^{(1)}\nonumber\\
    \alpha_{t, a} &= \text{softmax}\Big(\frac{(\matw^{(q)} \vece_t)^\top \cdot (\matw^{(k)} \vece_a)}{\sqrt{d/H}}\Big)
\end{align}
Our efficient self-attention involves embeddings $\vece_*^{(1)}, \vece_*^{(2)}, \vece_*^{(3)}$ for anchor nodes $v_a$ and vanilla nodes $v_t$. After each attention calculation, the results from multiple heads are concatenated, passed through a learnable linear layer, and connected with a residual connection. The parameters $\matw^{(q)}, \matw^{(k)}, \matw^{(v)}$ are the parameters of the attention layer. To reduce computational complexity, we employ a two-stage self-attention process. It transforms the $3\times B$-length sequence to a shorter $S$-length sequence and then reverses the process.

After the self-attention module, each layer of our scalable graph transformer includes a two-layer fully-connected block with residual connections, accompanied by two layer normalization modules. To ensure numerical stability, per-layer scaling is applied by element-wisely dividing embeddings by a selected constant $K=10$.

\subsubsection{Model Optimization}
\label{app:optimization}
To optimize our \model\ model, we utilize the masked autoencoding (MAE) training paradigm for self-supervised pre-training. Let's denote our \model\ as $f$, with trainable parameters $\param_f$, and a graph projection function $\phi$. The model is trained on a set of graphs $\graph_s$ with batch-specific labels $\bar{\sete}_s$. The objective of the generative SSL optimization is defined as follows:
\begin{align}
    \mathop{\arg\min}_{\param_f} \sum_{\graph_s}\sum_{\bar{\sete}_s\in\graph_s}
    \loss\Big(f(\graph_s-\bar{\sete}_s, \phi; \param_f), \nonumber\\
    \bar{\sete}_s\Big) + \lambda\cdot \|\param_f\|_\text{F}^2
\end{align}
Here, $\graph_s - \bar{\sete}_s$ represents the input graph $\graph_s$ with the label edge set $\bar{\sete}$ removed. All training graphs $\graph_s$ are jointly trained with random alternations. To enhance the model's adaptability to different graph projections $\phi$, we regenerate the projection function $\phi$ for each training graph $\graph_s$ every 10 training steps. $\lambda$ denotes the weight for $L_2$ regularization.

\subsection{Graph Generation Algorithm}
\label{app:generation}
\subsubsection{\bf Prompt Template and Examples of Generated Nodes}
\label{app:prompt}
\begin{figure}
    \centering
    \includegraphics[width=\columnwidth]{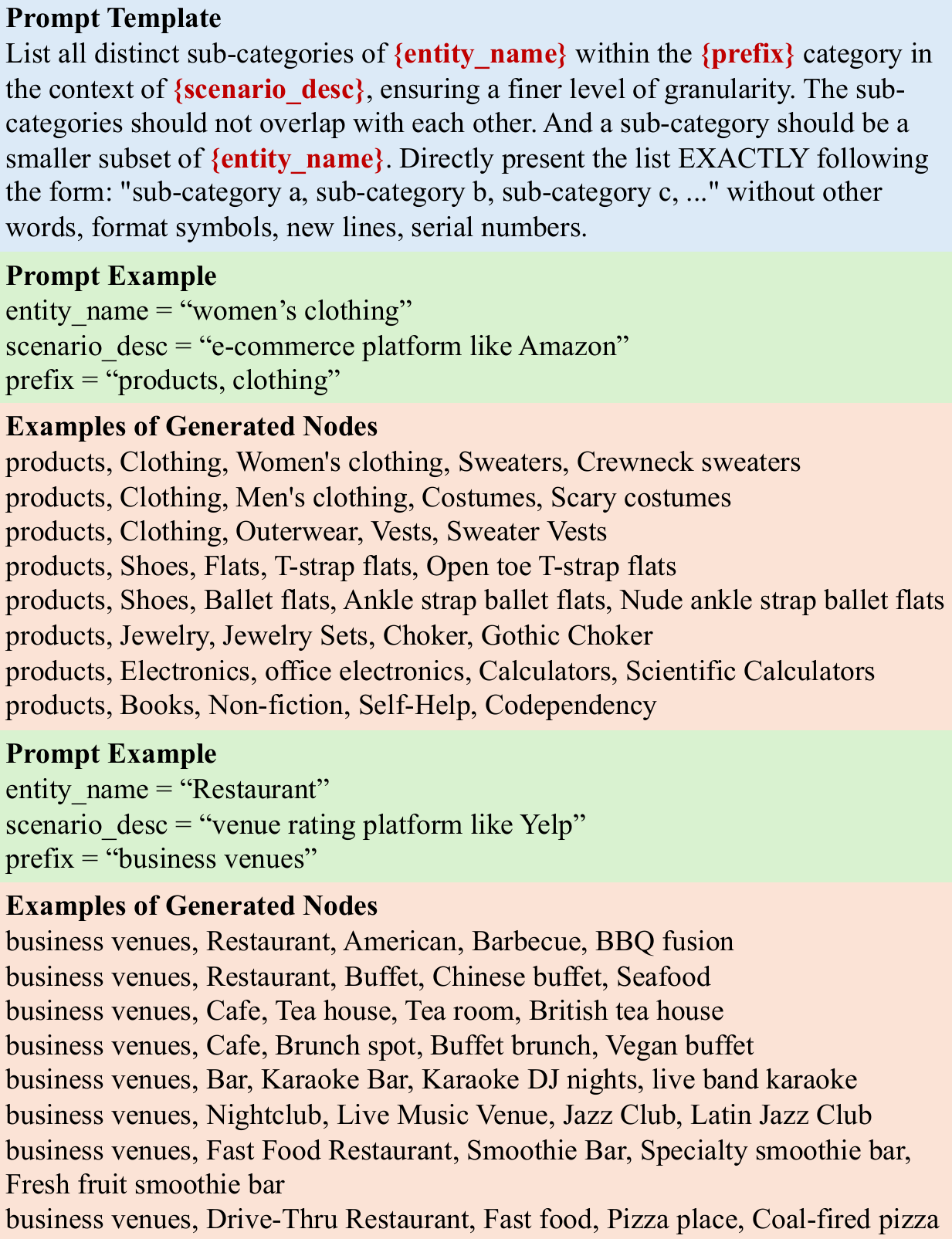}
    \caption{Prompt template and generation examples.}
    \label{fig:case_study}
\end{figure}

In this section, we present our prompt strategy for leveraging the Language Model (LLM) to divide general nodes into more fine-grained entities. Figure~\ref{fig:case_study} illustrates our prompt template, with key parameters highlighted in red. We provide concrete examples of prompt parameters and showcase the generation results for both the e-commerce scenario and the venue rating scenario.

\subsubsection{\bf Node Generation Algorithm}
\label{app:dfs}
We elaborate the process of our tree-of-prompt algorithm that traversing the vertex space for a specific application scenario, as in Algorithm~\ref{alg:dfs_alg}.
\begin{algorithm}[h]
    \caption{Tree-of-prompt algorithm for node generation.}
    \label{alg:dfs_alg}
    \KwIn{
        Name for the initial general node $v_0$ (\eg~'products'), text descriptions for the application scenario $S$ (\eg~'e-commerce platform like Amazon'), maximum depth $D$ of the prompt tree
    }
    \KwOut{
        Generated nodes $\hat{\mathcal{V}}$.
    }

    \SetKwFunction{FMain}{DivideNode}
    \SetKwProg{Fn}{Function}{:}{}
    \Fn{\FMain{$v$, $n$}}{
        \If{$n\geq D$}{\textbf{return} [$v$]}

        $\bar{\mathcal{V}}=\text{LLM}(v, S)$\\
        $\hat{\mathcal{V}}=[]$\\
        \ForEach{$v'\in\bar{\mathcal{V}}$}{
            $\hat{\mathcal{V}} +=$ \FMain{$v'$, $n+1$}
        }
        \textbf{return} $\hat{\mathcal{V}}$
    }
    \textbf{return} \FMain{$v_0$, $1$}
\end{algorithm}

\subsubsection{\bf Edge Generation Algorithm}
We elaborate our edge generation algorithm based on LLM-given node representations and the Gibbs sampling algorithm in Algorithm~\ref{alg:gibbs_alg}. Here we illustrate the case for generating person-entity relations, which is more complex compared to the entity-entity relation generation.
\label{app:gibbs}

\subsection{Experiments}
\begin{table}[h]
    \centering
    \small
    \setlength{\tabcolsep}{0.8mm}
    \caption{Statistics of experimental datasets.}
    \vspace{-0.12in}
    \label{tab:stat}
    \begin{tabular}{c|l|rrrc}
        \hline
        \multicolumn{2}{c|}{Dataset} & \# Node & \# Edge & \# Feat & \# Class\\
        \hline\hline
        \multirow{5}{*}{Link} & OGBL-ddi & 4,267 & 1,334,889 & 0 & \multirow{5}{*}{N/A}\\
        & OGBL-collab & 235,868 & 1,285,465 & 128 &\\
        & ML-1M & 9,746 & 720,152 & 0 & \\
        & ML-10M & 80,555 & 7,200,040 & 0 &\\
        & Amazon-book & 144,242 & 2,380,730 & 0 & \\
        \hline
        \multirow{3}{*}{Node} & Cora & 2,708 & 10,556 & 1433 & 7\\
        & Citeseer & 3,327 & 9,104 & 3,703 & 6\\
        & Pubmed & 19,717 & 88,648 & 500 & 3\\
        \hline
        \multirow{3}{*}{Gen.} & Gen0 & 46,861 & 454,276 & 0 & \multirow{3}{*}{N/A}\\
        & Gen1 & 51,061 & 268,007 & 0 &\\
        & Gen2 & 32,739 & 240,500 & 0 & \\
        \hline
        
    \end{tabular}
\end{table}

\subsubsection{\bf Experimental Datasets}
\label{app:dataset}
Our experimental datasets include 5 link prediction datasets and 3 node classification datasets. The data statistics are summarized in Table~\ref{tab:stat}.

\noindent\textbf{Link prediction datasets.} We employ five link prediction datasets from diverse application scenarios. The objective of these datasets is to predict the most likely connections for each node based on previous observations of node-wise interactions.\vspace{-0.08in}
\begin{itemize}[leftmargin=*]
    \item \textbf{OGBL-ddi}. This dataset is used for drug-drug interaction prediction. Each node represents a drug. The edges represent the combined effect of taking two drugs together, which differs significantly from taking them individually.\vspace{-0.12in}
    \item \textbf{OGBL-collab}. It is an academic social relation dataset. Its nodes represent scholars, and edges denote collaborations. Each node is combined with a 128-dimensional average word embedding calculated from the author's publications.\vspace{-0.12in}
    \item \textbf{Movielens-1M \& Movielens-10M}. These two datasets are both collected from the movie rating platform Movielens. The graphs are constructed by connecting users with the movies they have rated. The two datasets contain 1 million and 10 million rating records, respectively.\vspace{-0.12in}
    \item \textbf{Amazon-book}. This dataset contains review data from the Amazon platform. The nodes in the dataset represent users and books, while the edges denote the review records between them.
\end{itemize}

\noindent\textbf{Node classification datasets.} For the node classification task, we utilize three widely-used citation network datasets: \textbf{Cora}, \textbf{Citeseer}, and \textbf{Pubmed}. In these datasets, each node represents an academic paper, and an edge $(v_i, v_j)$ denotes a citation relation from paper node $v_i$ to $v_j$. The Cora and Citeseer datasets include binary bag-of-words vectors as node features, while the Pubmed dataset utilizes TF-IDF weighted word vectors as node features. The objective of these datasets is to classify each node into predefined paper categories based on the citation relations and node attributes.

\subsubsection{\bf Evaluation Protocol}
\label{app:evaluation_protocol}
This section includes detailed description for our cross-dataset zero-shot setting and the few-shot settings for baselines. It also introduces the evaluation metrics used in our experiments.

\noindent \textbf{Zero-shot setting}. For our \model, we utilize a zero-shot learning setting in which \model\ is not trained on any of these real-world datasets but is tested using the training set information as input information, including the graph structures, node features, and node labels in the training set.
To effectively generalize to unseen node labels in node classification with zero shot, taking inspiration from previous works~\cite{sun2022gppt}, we treat the label classes as new nodes and connect the vanilla nodes with training labels to the corresponding class nodes. This strategy removes the requirement for learning class-related parameters in the zero-shot learning setting. This enhancement is also applied to baselines methods.

\noindent \textbf{Few-shot setting}. Since most baselines perform poorly in the foregoing zero-shot setting, we evaluate them in the one-shot setting and five-shot setting.
In the node classification task, the \textit{k}-shot setting refers to preserving a maximum of \textit{k} training instances for each label class. For the link prediction task, the \textit{k}-shot training set contains at most \textit{k} links for each node.
Non-pretraining approaches such as MLP and GNNs are solely trained on the few-shot training set. On the other hand, baselines following the pretraining-and-tuning paradigm undergo pretraining and subsequent tuning on the few-shot set. In link prediction, they are pretrained on the same generated datasets as our \model. Model parameters that are not transferable across datasets are re-learned during the tuning phase. In node classification, these methods are pretrained on the graph of the target dataset and fine-tuned on the classification labels. In the test phase, all information in the training set is employed.

\noindent \textbf{Evaluation metrics}. In link prediction, we follow existing works~\cite{wei2024promptmm} to conduct the full-rank test for each node. To be specific, for each node, all nodes not connected to it in the training set are ranked by the model. The top-\textit{N} nodes are taken as positive predictions, and we calculate \textit{Recall@N} scores with $N=20,40$. In node classification, we employ the widely-used \textit{Accuracy} and \textit{Macro-F1} metrics~\cite{chen2022graph}.

\subsubsection{\bf Implementation Details}
\label{app:implementation}
We implemented our \model\ framework using PyTorch. The model employs the Adam optimizer with a learning rate of $1e-4$ or $5e-5$. The learnable parameters are initialized using the Xavier uniform initialization method. By default, the reported performance is achieved by \model\ with an embedding size of $d=1024$ and a maximum power order of $L=3$ for adjacency smoothing. The default scalable graph transformer utilizes $L'=3$ transformer layers, $H=4$ attention heads, and $S=256$ sampled anchor nodes. The training batch size, which is also used for token sequence sampling, is set as $B=1024$.

The reported results are obtained by pretraining our \model\ network using three generated datasets: Gen0, Gen1, and Gen2. The statistics of these datasets are presented in Table~\ref{tab:stat}.
We first generate the Gen0 dataset without injecting the graph's topological pattern. Subsequently, we generate Gen1 and Gen2 based on Gen0 by incorporating the graph pattern. In comparison to Gen1, the Gen2 dataset undergoes an additional densification process, where nodes with less than 10 edges are removed.
To acquire the nodes for the Gen0 dataset, we prompt the LLM to iterate through all products on an e-commerce platform, with a maximum generation depth of $5$. The Gibbs sampling algorithm is initialized with nodes having $6$ random edges. To ensure low overlap between consecutive samples, we introduce a separation of $1000$ sampling steps before generating each new sample. The dynamic probability normalization maintains the last $T'=5000$ sampling instances. The node locality incorporation involves using $7$ locality indices and $0.95$ decay rate.

The baseline methods are evaluated using their original code, or we closely follow the original code to implement them. Our implementations of the baselines are carefully aligned with the reported performance in their original evaluation settings. We employ grid search to optimize the hyperparameter settings for each baseline.

\subsubsection{\bf Baselines}
\label{app:baselines}
We give detailed descriptions for the baseline models in this section. 9 models from 4 different research lines are utilized in our evaluation.

\noindent \textbf{Graph-agnostic Approaches}.\vspace{-0.08in}
\begin{itemize}[leftmargin=*]
    \item \textbf{MF}. This is the matrix factorization approach which learns node embeddings to reconstruct the observed adjacency matrix. For the node classification task, we adapt it to learn embedding vectors for each node to predict node labels.\vspace{-0.12in}
    \item \textbf{MLP}. This baseline utilizes a multi-layer perceptron to extract deep features individually for each node. For datasets without node attributes, this baseline learns initial node embeddings.
\end{itemize}

\noindent \textbf{Non-pretraining Graph Neural Networks}.\vspace{-0.08in}
\begin{itemize}[leftmargin=*]
    \item \textbf{GCN}~\cite{kipf2016semi}. This approach utilizes iterative graph convolutional operators to extract the high-order topological information.\vspace{-0.12in}
    \item \textbf{GAT}~\cite{velivckovic2017graph}. This graph attention network learns weights for node-wise connections using the attention mechanism, to facilitate adaptive graph information propagation.\vspace{-0.12in}
    \item \textbf{GIN}~\cite{xu2018powerful}. This method enhances the representation power of GNNs by employing a distinct graph encoding method that emphasizes the discrimination of non-isomorphic structures.
\end{itemize}

\noindent \textbf{Graph Pre-training Models}.\vspace{-0.08in}
\begin{itemize}[leftmargin=*]
    \item \textbf{GraphCL}~\cite{zhu2021graph}. This baseline method utilizes pre-training of graph models through the application of a self-discriminative contrastive learning task on learned node embeddings. It incorporates various graph augmentation techniques such as node drop, edge permutation, random walk, and feature masking.\vspace{-0.12in}
    \item \textbf{DGI}~\cite{velivckovic2018deep}. This method introduces a self-supervised pre-training task that aims to maximize the mutual information between the local view and the global view.
\end{itemize}

\noindent \textbf{Graph Prompt Tuning Methods}.\vspace{-0.08in}
\begin{itemize}[leftmargin=*]
    \item \textbf{GraphPrompt}~\cite{liu2023graphprompt}. This work presents a unified framework for pre-training and prompt tuning of graph models. It introduces a learnable prompt layer that automatically identifies crucial information in the pre-trained model to facilitate downstream tasks.\vspace{-0.12in}
    \item \textbf{GPF}~\cite{fang2022universal}. This is a universal graph prompt tuning framework designed for various graph pre-training strategies. It introduces two versions of a learnable graph prompt layer.
\end{itemize}

\subsubsection{\bf Alternative Projection Methods (RQ2)}
\label{app:proj_ablation}
\begin{itemize}[leftmargin=*]
    \item \textbf{One-hot encoding}. This graph projection strategy employs a large table of low-dimensional embeddings for node ids, where nodes with the same index from different datasets are directly mapped to the same embedding vector. Specifically, we utilize 100,000 independent embedding vectors. For datasets with more nodes, we use the remainder of 100,000 dividing node indices. \vspace{-0.12in}
    
    \item \textbf{Degree embeddings}. Using degree embeddings for node representation is a commonly used strategy for non-attributed graphs. Each degree number is assigned an independent learnable embedding vector, and each node is initially represented by its degree representation. \vspace{-0.12in}
    
    \item \textbf{Random projection}. In this approach, a random representation vector is assigned to each node, sampled from a uniform distribution. With a sufficiently large representation space, this method aims to approximately distribute nodes with equal distances from one another. As a result, this projection method does not rely on any specific assumptions about the node distribution. This characteristic allows it to outperform the other two strategies, which are based on certain assumptions and are thus more prone to overfitting the pre-training dataset.

\end{itemize}

\subsubsection{\bf End-to-end Training (RQ6)}
\label{app:end_to_end_comparison}
\begin{table}[]
    \centering
    \small
    \setlength{\tabcolsep}{0.6mm}
    \caption{Performance comparison with models trained on few-shot datasets, in terms of Recall@20 (\%).}
    \vspace{-0.12in}
    \label{tab:fewshot}
    \begin{tabular}{l|cc|cc|cc|cc}
        \hline
        Model & \multicolumn{2}{c|}{GCN} & \multicolumn{2}{c|}{GAT} & \multicolumn{2}{c|}{GIN} & \multicolumn{2}{c}{\model}\\
        shot & 1 & 5 & 1 & 5 & 1 & 5 & 1 & 5\\
        \hline
        \hline
        ddi & 2.79 & 7.05 & 5.80 & 7.11 & 5.30 & 7.35 & \textbf{7.93} & \textbf{8.60}\\
        ML1M & 4.32 & 10.54 & 2.45 & 15.06 & 4.66 & 14.58 & \textbf{16.58} & \textbf{19.19}\\
        Amazon & 0.96 & 2.51 & 0.47 & 2.28 & 0.69 & 2.52 & \textbf{2.96} & \textbf{3.10}\\
        \hline
    \end{tabular}
\end{table}

We compare our \model\ with other graph encoding methods in the supervised learning setting. The models are trained using the 1-shot and 5-shot training sets from OGBL-ddi, ML-1M, and Amazon-book, and then tested on the corresponding test set. Without pre-training, this experiment aims to examine the modeling capacity for different graph neural architectures. From the results shown in Table~\ref{tab:fewshot}, we draw the following conclusions:\vspace{-0.08in}
\begin{itemize}[leftmargin=*]
    \item \textbf{Superior modeling capabilities of \model}. Our \model\ achieves best performance on all tested datasets, demonstrating the superior graph learning ability for \model. We attribute this superiority to the precise preservation of structural information by our graph tokenization module, and the strength of our scalable graph transformer in learning global relations.

    \item \textbf{Robustness of \model}. We notice that our \model\ exhibits less performance degradation on the more sparse 1-shot datasets. This demonstrates the inherent robustness of \model's model architecture. Such robustness can be ascribed to the effectiveness of the fast topology projection, which effectively captures key graph structures even without sufficient training.
\end{itemize}

\begin{algorithm}[h]
    \caption{Edge generation algorithm.}
    \label{alg:gibbs_alg}
    \KwIn{
        Node embedding table $\textbf{H}$ given by the LLM, node set $\setv$,
        maximum locality index $N$, locality decay factor $\alpha$, dynamic probability normalization range $T'$, number of sampling steps to draw a new sample $T_0$, number of initial sampling steps to skip for data quality $T_1$, number of sampling steps to shift current locality index $T_2$, maximum sampling steps $T_\text{max}$.
    }
    \KwOut{
        List of interactions $\mathcal{I}$.
    }
    Draw a random interaction sample $\veca^0$\\
    Initialize current locality index $n=0$\\
    Initialize the pool for dynamic probability $\mathcal{P}=[]$\\
    Initialize $\mathcal{I}=[]$
    \For{$t=1$ to $T_\text{max}$}{
        \If{$t~\text{mod}~T_2 ==0$}{
            $n=(n+1)~\text{mod}~N$
        }
        $i = t~\text{mod}~|\setv|$\\
        $p = \sum_{v_{i}} a^t_i (\vech_i / \|\veca^t\|_0)^\top \cdot \vech_{t'}$\\
        $\mathcal{P}+=[p]$\\
        \If{$|\mathcal{P}|>T'$}{
            $\mathcal{P}=\mathcal{P}[-T':]$
        }
        $\mu= \text{mean}(\mathcal{P})$, $\sigma=\text{std}(\mathcal{P})$\\
        $\bar{p}=(p-\mu)/(4\sigma)$\\
        $\hat{p}=\bar{p} \cdot \alpha^{|n-n_i|}$\\
        Decide if $\veca^t\oplus v_{t'}$ is accepted accordding to $\hat{p}$\\
        \If{$t\geq T_1$ and $t~\text{mod}~T_0==0$}{
            $\mathcal{I}+=[\veca^t\oplus v_{t'}]$
        }
    }
    \textbf{return} $\mathcal{I}$
\end{algorithm}